\documentclass[pdflatex,sn-mathphys-num]{sn-jnl}

\usepackage{graphicx}%
\usepackage{multirow}%
\usepackage{amsmath,amssymb,amsfonts}%
\usepackage{amsthm}%
\usepackage{mathrsfs}%
\usepackage[title]{appendix}%
\usepackage{xcolor}%
\usepackage{textcomp}%
\usepackage{manyfoot}%
\usepackage{booktabs}%
\usepackage{algorithm}%
\usepackage{algorithmicx}%
\usepackage{algpseudocode}%
\usepackage{listings}%
\usepackage{hyperref}

\theoremstyle{thmstyleone}%
\theoremstyle{thmstyletwo}%

\theoremstyle{thmstylethree}%

\begin{document}

\title[Article Title]{Exploiting ChatGPT for Diagnosing Autism-Associated Language Disorders and Identifying Distinct Features}


\author[1]{\fnm{Chuanbo} \sur{Hu}}\email{chu3@albany.edu}

\author[1,2]{\fnm{Wenqi} \sur{Li}}\email{wl00021@mix.wvu.edu}

\author[2]{\fnm{Mindi} \sur{Ruan}}\email{mr0114@mix.wvu.edu}

\author[3]{\fnm{Xiangxu} \sur{Yu}}\email{xiangxu@wustl.edu}

\author[4]{\fnm{Shalaka} \sur{Deshpande}}\email{mr0114@mix.wvu.edu}

\author[5]{\fnm{Lynn K.} \sur{Paul}}\email{lkpaul@hss.caltech.edu}

\author[3]{\fnm{Shuo} \sur{Wang}}\email{lkpaul@hss.caltech.edu}

\author[1]{\fnm{Xin} \sur{Li}}\email{xli48@albany.edu}


\affil*[1]{\orgdiv{Department of Computer Science}, \orgname{University at Albany}, \city{Albany}, \postcode{12222}, \state{NY}, \country{USA}}

\affil[2]{\orgdiv{Lane Department of Computer Science and Electrical Engineering}, \orgname{West Virginia University},  \city{Morgantown}, \postcode{26506}, \state{WV}, \country{USA}}

\affil[3]{\orgdiv{Department of Radiology}, \orgname{Washington University in St. Louis},  \city{St. Louis}, \postcode{63110}, \state{MO}, \country{USA}}

\affil[4]{\orgname{Camas High School},  \city{Camas}, \postcode{98607}, \state{WA}, \country{USA}}

\affil[5]{\orgdiv{Humanities and Social Sciences}, \orgname{California Institute of Technology}, \city{Pasadena}, \postcode{91125}, \state{CA}, \country{USA}}


\abstract{Diagnosing language disorders associated with autism is a complex challenge, often hampered by the subjective nature and variability of traditional assessment methods. Traditional diagnostic methods not only require intensive human effort but also often result in delayed interventions due to their lack of speed and precision. In this study, we explored the application of ChatGPT, a state-of-the-art large language model, to overcome these obstacles by enhancing sensitivity and profiling linguistic features for autism diagnosis. This research utilizes ChatGPT's natural language processing capabilities to simplify and improve the diagnostic process, focusing on identifying autism-related language patterns. Specifically, we compared ChatGPT's performance with that of conventional supervised learning models, including BERT, a model acclaimed for its effectiveness in various natural language processing tasks. We showed that ChatGPT substantially outperformed these models, achieving over 10\% improvement in both sensitivity and positive predictive value, in a zero-shot learning configuration. The findings underscore the model’s potential as a diagnostic tool, combining accuracy and applicability. We identified ten key features of autism-associated language disorders across scenarios. Features such as echolalia, pronoun reversal, and atypical language usage play a critical role in diagnosing ASD and informing tailored treatment plans. Together, our findings advocate for adopting sophisticated AI tools like ChatGPT in clinical settings to assess and diagnose developmental disorders. Our approach promises enhanced diagnostic precision and supports personalized medicine, potentially transforming the evaluation landscape for autism and similar neurological conditions.}




\maketitle

\section{Introduction}\label{sec1}

Autism spectrum disorder (ASD) is a developmental condition marked by difficulties in social interaction, restricted interests, and repetitive behaviors. \cite{leekam2011restricted,lord2018autism,lord2020autism}. The spectrum of ASD symptoms is broad, with communication difficulties often standing out as the most significant and impacting aspects of the disorder \cite{lauritsen2013autism,baird2016social,nazeer2012autism}. These symptoms vary across age groups. In adults, communication difficulties are especially pronounced, significantly affecting social integration and personal development \cite{sperry2005perceptions,velikonja2019patterns}. Identifying these communication issues is crucial for effective intervention.

Language disorders in adults with ASD include a wide range of issues, from the absence of speech to subtle impairments such as echolalia (repetitive use of phrases or sounds), pronoun reversal, and pragmatic language impairments \cite{rapin2003update,magiati2014cognitive,whitehouse2009adult}. These disorders can hinder effective communication and pose significant challenges in social and occupational settings. Accurate identification of language anomalies is essential for timely interventions.

While the Autism Diagnostic Observation Schedule, Second Edition (ADOS-2) \cite{lord2012autism} is a gold standard for ASD diagnosis across various age groups. The ADOS-2 involves structured, standardized observation sessions, typically lasting around one hour, during which an examiner engages the patient in a series of conversational and interactive tasks. These sessions are designed to assess key areas of communication, social interaction, and behavior. For adults, the examiner analyzes the patient's responses, verbal and non-verbal communication patterns, and social cues to identify markers of ASD. The diagnostic process relies on the examiner's ability to interpret subtle interaction dynamics and communication behaviors, making the method highly reliant on clinical expertise. As a result, the approach can be both subjective and resource-intensive, posing challenges for scalability and consistency in assessments.

To address these challenges, researchers have turned to machine learning (ML) as a complementary or alternative diagnostic tool. ML models leverage multimodal data encompassing speech, text, video, and image to detect ASD-related markers with greater automation and precision \cite{hu2024exploring,mukherjee2023detection,yu2024video,ruan2023video,ruan2024can,zhang2022discriminative}. For example: A contrastive learning framework for decision tree-based action classification leverages adjacency matrices and skeleton graphs to model periodicity and symmetry, enabling robust recognition of human interactions with potential in video-based ASD diagnosis \cite{ruan2023video}. A few-shot learning framework leveraging facial dynamics and scene-level fusion analyzes hour-long ADOS videos, classifying individuals into Autism, Autism Spectrum, and Non-Spectrum with 91.72\% accuracy while highlighting the value of specific interview scenes \cite{zhang2022discriminative}. While ML approaches have shown promise, they face significant challenges: 1) \textbf{High Data Requirements:} Training effective ML models requires large, labeled datasets. However, collecting such data is expensive, as it involves long, structured interactions between examiners and patients \cite{ellis2010early}. 2) \textbf{Lack of Explainability:} Many ML models operate as “black boxes,” offering little insight into how predictions are made. This reduces their trustworthiness in clinical applications.

Large language models (LLMs), such as GPT \cite{brown2020language}, represent a paradigm shift in ASD diagnosis. Unlike traditional ML models, LLMs are pre-trained on massive corpora of diverse text data, enabling them to generalize across tasks with minimal task-specific training. Their applications in healthcare, including mental health diagnostics and personalized medicine, are gaining attention \cite{thirunavukarasu2023large,yang2022large}. LLMs offer several advantages over traditional ML approaches: 1) \textbf{Zero-Shot and Few-Shot Learning Capabilities:} LLMs excel in low-data scenarios, leveraging pre-trained knowledge to identify patterns and make predictions without requiring extensive labeled datasets \cite{wei2024chatie}. This is particularly valuable for ASD diagnosis, where data collection is costly and labor-intensive; 2) \textbf{Rich Linguistic and Contextual Knowledge:} LLMs can capture nuanced language patterns, such as echolalia and atypical prosody, directly from text data. Unlike conventional NLP models, they require minimal feature engineering and can adapt to complex linguistic contexts \cite{liu2023summary}; 3) \textbf{Explainability:} LLMs can generate detailed, human-like explanations for their predictions, enhancing transparency and trust in clinical settings \cite{quttainah2024cost}.

This study leverages ChatGPT, a state-of-the-art LLM, to enhance the sensitivity and efficiency of ASD diagnosis through its ability to process nuanced linguistic patterns. By analyzing features such as echolalia, pronoun reversals, and atypical prosody, ChatGPT provides critical insights into key indicators of ASD. Its zero-shot and few-shot learning capabilities effectively address the challenges of limited ASD-specific datasets, reducing reliance on costly and labor-intensive data collection. Furthermore, ChatGPT’s natural language understanding enables it to analyze examiner-patient dialogues in a conversational context, capturing subtle language behaviors essential for diagnosis. The model’s human-like explanations further support clinicians, fostering trust and transparency in diagnostic outcomes. By integrating domain-specific knowledge and multimodal data such as gaze patterns and non-verbal behaviors, ChatGPT could pave the way for a comprehensive AI-driven framework for ASD language assessment. Despite their potential, applying LLMs in ASD diagnosis remains an emerging field with following motivations:
\begin{itemize}
    \item \textbf{Advancing Diagnostic Sensitivity and Efficiency.}
    
    Diagnosing autism-related language disorders is challenging due to subtle and variable symptoms. This study leverages ChatGPT to achieve substantial improvements in sensitivity (recall), ensuring better detection of language deficits while streamlining the diagnostic process for timely interventions.

    \item \textbf{Identification of Specific Linguistic Features.}
    
    Current methods for identifying language disorders in autistic individuals often rely on broad assessments that miss specific features. ChatGPT’s advanced pattern recognition refines this process by identifying key linguistic markers like echolalia, pronoun reversal, and atypical language use. These insights are crucial for developing personalized treatment plans.

    \item \textbf{Enabling Personalized Therapeutic Strategies.}

    By combining high sensitivity and precision, ChatGPT facilitates targeted therapeutic approaches for ASD, addressing individual patient needs. This framework also opens avenues for extending AI-driven diagnostic tools to other cognitive and developmental disorders.

\end{itemize}

Building on the limitations of traditional diagnostic tools, this study leverages ChatGPT's language processing capabilities to address these challenges. By comparing its performance with existing supervised learning models, we aim to highlight the potential of LLMs in accurately identifying language deficits associated with ASD. This research integrates advanced AI technologies with domain-specific knowledge to enhance diagnostic efficiency while identifying distinct linguistic features indicative of ASD. The following sections outline the datasets, experimental setup, and model comparisons to substantiate these claims.

\section{Results}

\subsection{Experimental Dataset}

\subsubsection{ADOS-2} 
The ADOS-2 \cite{lord1999ados,american2013diagnostic} is an update and extension of the original ADOS, which is a standardized diagnostic tool for ASD. The ADOS assesses communication, social interaction, play, and restricted and repetitive behaviors. It provides a series of structured and semi-structured tasks that involve social interactions between the examiner and the person being assessed. Module 4 of the ADOS-2 is designed for verbally fluent adolescents and adults (see Table \ref{tab:scenarios} for description of tasks). In addition, Module 4 of the ADOS-2 organizes observations into five main areas, assessing various aspects of interaction and communication critical for diagnosing ASD in verbally fluent adolescents and adults. Table \ref{tab:score} provides a summary of these categories, including the specific items they encompass and their respective descriptions: each participating in 15 different scenarios (see Table \ref{tab:scenarios}) designed to elicit communicative and social responses that are indicators of ASD. The scenarios were structured to cover a comprehensive range of social interactions and communicative behaviors.

\begin{table}
\centering
\small
\begin{tabular}{|c|p{3cm}|p{8.4cm}|}
\hline
\textbf{Scenario} & \textbf{Name} & \textbf{Explanation} \\ \hline
\(S_1\) & Construction Task & Involves the participant engaging in a task that requires constructing or assembling a set structure, testing spatial and motor skills rather than communicative abilities. \\ \hline
\(S_2\) & Telling a Story from a Book & Primarily a monologic task where the participant recounts a story from a book, differing from spontaneous dialogic interactions. \\ \hline
\(S_3\) & Description of a Picture & Participants describe a picture, testing their ability to interpret visual information and articulate a coherent description. \\ \hline
\(S_4\) & Conversation and Reporting & Focuses on the ability to engage in back-and-forth conversation and to report on past events. \\ \hline
\(S_5\) & Current Work and School & Discusses participants' current educational and occupational engagements. \\ \hline
\(S_6\) & Social Difficulties and Annoyance & Elicits experiences of social challenges and annoyances. \\ \hline
\(S_7\) & Emotions & Requires participants to express and identify emotions. \\ \hline
\(S_8\) & Demonstration Task & Requires the participant to demonstrate how to use an item or explain a process, which does not involve interactive communication with an examiner. \\ \hline
\(S_9\) & Cartoons & Involves interpreting sequences and explaining cartoon strips. \\ \hline
\(S_{10}\) & Break & A pause or intermission in the assessment, involving no communicative or cognitive tasks. \\ \hline
\(S_{11}\) & Daily Living & Covers daily routines and personal care tasks. \\ \hline
\(S_{12}\) & Friends, Relationships, and Marriage & Discusses personal relationships and social norms regarding friendships and marital status. \\ \hline
\(S_{13}\) & Loneliness & Addresses feelings and situations of loneliness and isolation. \\ \hline
\(S_{14}\) & Plans and Hopes & Involves discussing future aspirations and plans. \\ \hline
\(S_{15}\) & Creating a Story & Tests creative storytelling abilities in an unstructured task. \\ \hline
\end{tabular}
\caption{Overview of Scenario Tasks in ADOS-2 Module 4 Diagnosing process}
\label{tab:scenarios}
\end{table}

\begin{table}[h]
\centering
\caption{Detailed Assessment Categories for the ADOS-2 Module 4 Observations}
\label{tab:score}
\begin{tabular}{|c|p{2cm}|p{1cm}|p{7.5cm}|}
\hline
\textbf{Class} & \textbf{Name} & \textbf{Items} & \textbf{Description} \\
\hline
A & Language and Communication & A1 $\sim$ A10 & Assesses the ability to use speech and gestures in communication effectively, evaluating the clarity, coherence, and appropriateness of language used in social interactions. \\
\hline
B & Reciprocal Social Interaction & B1 $\sim$ B13 & Focuses on non-verbal and verbal behaviors used in social interactions, including eye contact, facial expressions, body postures, and the quality of speech interactions. \\
\hline
C & Imagination $/$ Creativity & C1 & Evaluates the subject's ability to use imagination and creativity in their expressions and thoughts, such as storytelling or creating novel responses to social scenarios. \\
\hline
D & Stereotyped Behaviors and Restricted Interests & D1 $\sim$ D5 & Includes specific behaviors that are repetitive, restricted, and stereotyped. This category assesses the frequency and intensity of these behaviors as indicators of ASD. \\
\hline
E & Other Abnormal Behaviors & E1 $\sim$ E3 & Observes behaviors that are typically considered abnormal, such as overactivity, anxiety, and emotional responses that are inconsistent with the normative context. \\
\hline
\end{tabular}
\end{table}

\begin{table}[h]
\centering
\caption{A4 Score: Stereotyped/Idiosyncratic Use of Words or Phrases}
\begin{tabular}{|c|p{12cm}|}
\hline
\textbf{Score} & \textbf{Description} \\ \hline
0 & Rarely or never uses stereotyped or idiosyncratic words or phrases. The subject demonstrates typical language use without noticeable patterns of repetition or unusual phrasing. \\ \hline
1 & Uses words or phrases that are more repetitive or formal compared to most individuals at a similar level of expressive language, though not obviously odd. This category also includes occasional stereotyped utterances or odd use of words or phrases, while still showing substantial spontaneous and flexible language use. \\ \hline
2 & Often uses stereotyped utterances or odd words or phrases, with some other spontaneous language. The subject displays a noticeable pattern of repetitive or unusual phrasing that stands out in conversation. \\ \hline
3 & Frequently uses odd or stereotyped speech and rarely exhibits non-stereotyped spontaneous speech. The language is predominantly characterized by repetitive, formal, or idiosyncratic expressions, with very little flexibility or spontaneity. \\ \hline
\end{tabular}
\label{table:A4Score}
\end{table}

In language-based diagnostics, the \textbf{A4 score} — part of the Stereotyped Behaviors and Restricted Interests category — becomes particularly relevant. A4 score can be evaluated using the mechanism described in Table \ref{table:A4Score}:

The A4 score assesses the use of stereotyped language, which is a critical indicator of ASD. A higher A4 score suggests a more frequent use of stereotyped or idiosyncratic speech, aiding in the diagnosis of ASD with higher specificity and sensitivity.

\subsubsection{Autism-Associated Language Disorders features}

Based on the domain knowledge of professional ADOS-2 examiners, ten specific features of language deficits related to ASD have been identified. These features reflect various unconventional use patterns of language that can signify underlying social communication issues. Table \ref{tab:3} is a detailed description of these features:

\begin{table}[h]
\centering

\begin{tabular}{|c|l|p{9cm}|} 
\hline
\textbf{F} & \textbf{Name} & \textbf{Explanation} \\ \hline
\(F_1\) & \begin{tabular}[c]{@{}l@{}}Echoic\\ Repetition\end{tabular} & The individual mimics verbatim what has been said by others, including the examiner, or recites phrases from external sources like advertisements or movie scripts, showing a delayed echo response. \\ \hline
\(F_2\) & \begin{tabular}[c]{@{}l@{}}Unconventional\\ Content\end{tabular} & The speech contains peculiarly chosen content or contextually odd phrasing, such as using 'unfreshness through household' for lack of novelty, 'mideast' instead of 'midwest' for U.S. states, or describing entry into a building as 'through various apertures'. \\ \hline
\(F_3\) & \begin{tabular}[c]{@{}l@{}}Pronoun\\ Displacement\end{tabular} & Incorrectly substitutes personal pronouns, using 'you' in place of 'I', or refers to themselves in the third person, either by pronouns like 'he/she' or by their own name. \\ \hline
\(F_4\) & \begin{tabular}[c]{@{}l@{}}Incongruous \\ Humor Timing\end{tabular} & Incorporates humorous or comedic expressions inappropriately during discussions meant to be serious, showing a misalignment between the content's emotional tone and the context. \\ \hline
\(F_5\) & \begin{tabular}[c]{@{}l@{}}Formalistic\\ Language Use\end{tabular} & Employs an overly formal or archaic language style that seems lifted from written texts, legal documents, or old literature, rather than engaging in conversational speech. Examples include elaborate ways of expressing simple ideas or feelings. \\ \hline
\(F_6\) & \begin{tabular}[c]{@{}l@{}}Superfluous \\Phrase Attachment\end{tabular} & Attaches redundant phrases or filler expressions to their speech without contributing any substantive meaning or context, such as 'you know what I mean' or 'as they say,' indicating a habit rather than intentional emphasis. \\ \hline
\(F_7\) & \begin{tabular}[c]{@{}l@{}}Excessive Social\\ Phrasing\end{tabular} & Utilizes conventional social expressions excessively or inappropriately, responding with phrases like 'oh, thank you' in contexts where it does not fit or preempting social gestures not yet performed by the interlocutor. \\ \hline
\(F_8\) & \begin{tabular}[c]{@{}l@{}}Monotone Social\\ Expression\end{tabular} & Reiterates social phrases with an unchanged, monotonous intonation, indicating a lack of genuine emotional engagement or variability in social interactions. \\ \hline
\(F_9\) & \begin{tabular}[c]{@{}l@{}}Stereotyped Media\\ Quoting\end{tabular} & Quotes lines from commercials, movies, or TV shows in a highly stereotypical manner, employing a canned intonation that mimics the original source closely, suggesting a reliance on external media for verbal expressions. \\ \hline
\(F_{10}\) & \begin{tabular}[c]{@{}l@{}}Clichéd Verbal\\ Substitutions\end{tabular} & Resorts to well-known sayings or clichés in lieu of engaging in direct conversational responses, using phrases like 'circle of life' or 'ready to roll' as stand-ins for more personalized communication. \\ \hline
\end{tabular}
\caption{Descriptive Analysis of Unconventional Language Disorder Patterns}
\label{tab:3}
\end{table}

\subsubsection{ADOS-2 Interview Audios}

For the purpose of this study, we utilized the Caltech ADOS Audio Dataset \cite{zhang2022discriminative}, which comprises audio recordings from diagnostic interviews conducted under the ADOS-2, Module 4. This dataset \footnote{\href{https://github.com/cbhu523/chatgpt_ASD_diagnosis}{Caltech Dataset}} involved 33 participants, aged 16 to 37 years (mean age:24.32), with a composition of 26 males and 7 females, predominantly right-handed (n=31). A subset of 9 participants underwent two ADOS interviews approximately six months apart, resulting in a total of 42 audio recordings spanning 3,165 minutes. Among these, 14 audio recordings have an A4 score of 0, 23 have an A4 score of 1, 5 have an A4 score of 2, and none have an A4 score of 3. Each recording was segmented into 15 sub-videos based on scene timestamps, producing 292 sub-videos with an average duration of 334 seconds. All were diagnosed with ASD based on the ADOS-2, confirmed through expert clinical evaluation. In addition, Each audio participated in 15 different scenarios that are designed to elicit social and communicative behaviors characteristic of individuals on the autism spectrum. Our research focused exclusively on scenarios involving direct dialogue between the examiner and the patient, as this interaction is critical for analyzing language disorders in autism. Therefore, we excluded scenarios like the \(S_1\) Construction Task and \(S_8\) Demonstration Task, which involve non-verbal or explanatory tasks, lacking the necessary dialogue. The \(S_2\) Tell a Story scenario was also excluded because it involves monologic speech without interactive exchange. Additionally, the \(S_{10}\) Break scenario does not include any relevant communicative or cognitive tasks. These exclusions ensured that our analysis remained focused on interactions that are directly relevant to diagnosing language-related symptoms in ASD. Out of these scenarios, we specifically selected 11 that focus on social language interactions between the examiner and the patient. These selected scenarios provide a concentrated dataset to analyze social communicative exchanges, crucial for identifying language-related symptoms of ASD.


Among the selected scenarios, particular emphasis was placed on analyzing the linguistic feature of "Stereotyped/Idiosyncratic Use of Words or Phrases," which is central to understanding language disorders that may or may not be present in individuals with autism. A4 score details are presented in Table \ref{table:A4Score}. In our dataset, 13 subjects had a score of 0, indicating minimal or no use of stereotyped language, while twenty-seven had a score of 1, showing mild repetitive or formal use, and four had a score of 2, indicating frequent use of stereotyped or odd phrases. To simplify our binary classification task, we combined subjects with scores 1 and 2 into a single category (Category 1) and grouped those with a score of 0 into Category 0. This approach was necessary due to the limited number of subjects with more severe symptoms, as our sample primarily consists of adults with less pronounced symptoms. We also recognize the suggestion to include a more diverse set of dialogues, such as those from non-autistic populations, in future analyses to broaden the perspective on language use and disorders.

\subsection{Experimental Setup}
\textbf{Speaker Diarization Baseline Models.} 
To provide a comprehensive comparison with our Google-based speaker diarization approach, we evaluated two baseline models: Microsoft Azure Speaker Diarization and Pyannote speaker-diarization-3.1. The Microsoft Azure system was configured with SpeakerDiarizationWithTranscription mode, a maximum of 2 speakers, and output in detailed mode, providing timestamps, speaker attribution, and confidence scores. The input audio files were preprocessed to single-channel WAV format with a sampling rate of 16 kHz. In contrast, Pyannote’s speaker-diarization-3.1 pipeline, leveraging pre-trained pyannote/segmentation-3.0 and clustering algorithms, was configured with a minimum of 2 speakers and a maximum of 5 speakers, along with tuned overlap handling and segmentation thresholds. Both baseline models served to benchmark the diarization accuracy and robustness of our Google-based implementation, highlighting differences in handling overlapping speech and speaker variability.

\textbf{Text Classification Baseline Models.} The models we compared include XLNet, ALBERT, DistilBERT, RoBERTa, and BERT, all of which have a token limit of 512. Each scenario from a subject in our dataset spans 8-10 minutes of dialogue. In most cases, the token count stays within this range, with only a few samples exceeding the 512-token limit. To ensure a consistent and fair comparison, we standardized each sample by keeping only the first 512 tokens for those that exceeded the limit, while samples with fewer than 512 tokens were left unchanged. We evaluated these baseline models using 5-fold cross-validation to ensure robust performance assessment. This involved splitting the dataset into five subsets, with each subset serving as the validation set once while the others were used for training. The final performance metrics were averaged across all five folds to account for variations due to data splits and provide a comprehensive evaluation.

\textbf{Evaluation Metrics.} The effectiveness of the ChatGPT model was evaluated using several metrics: Accuracy: The proportion of total diagnoses that were correctly identified by the model. Positive Predictive Value (PPV): The ratio of correct positive observations to the total predicted positives. sensitivity: The ratio of correct positive observations to the actual positives in the data. F1 Score: The harmonic mean of PPV and sensitivity, providing a single metric to assess the balance between PPV and sensitivity.

\textbf{Hardware.} All baseline models were trained on a workstation equipped with two NVIDIA 3090 Ti GPUs, providing a total of 48 GB of VRAM. This setup ensured sufficient computational resources for processing datasets and fine-tuning the models efficiently. The hardware allowed for accelerated training times and facilitated the handling of memory-intensive tasks such as tokenization and batch processing, which were critical for achieving consistent results across all models.

\subsection{Comparison of Language Deficit Diagnosis}
\subsubsection{Evaluation of LLM Performance}

Due to data limitations, cases with an A4 score of 3, representing extremely severe conditions, are not included in the dataset, and cases with an A4 score of 2 are extremely rare (42 audio recordings, 14 A4 = 0, 23 A4 = 1, 5 A4 = 2, and 0 A4 = 3.). Therefore, we categorized A4 $>$ 0 as Label = 1 and A4 = 0 as Label = 0 to explore whether large models can assist in diagnosing patients and identifying the presence of social language disorders. To compare the performance of different GPT versions, our backbone models were evaluated based on the API versions GPT-3.5-turbo and GPT-4. Table \ref{tab:1} summarizes the results of this comparison.

\begin{table}[tb]
\centering
\caption{Performance Metrics of Different Models}
\label{tab:1}
\begin{tabular}{|l|c|c|c|c|}
\hline
\textbf{Model} & \textbf{Accuracy} & \textbf{PPV} & \textbf{Sensitivity} & \textbf{F1 Score} \\ 
\hline
XLNet \cite{yang2019xlnet}            & 58.76\%           & 54.38\%            & 58.76\%         &  56.07\%           \\ 
ALBERT \cite{lan2019albert}            & 69.07\%           & 47.71\%            & 69.07\%         & 56.44\%           \\ 
DistilBERT \cite{sanh2019distilbert}      & 58.76\%           & 51.85\%            &  58.76\%         & 54.44\%           \\ 
RoBERTa \cite{liu2019roberta}        & 57.73\%           & 58.12\%            & 57.73\%         & 57.92\%           \\ 
BERT \cite{devlin2018bert}            & 63.92\%           & 60.86\%            & 63.92\%         & 61.87\%           \\ 
\hline
GPT 4o - based (Ours)            & 67.05\%           &  70.76\%            &  90.51\%         & 79.43\%           \\ 
GPT 3.5 - based (Ours)            & \textbf{69.14\%}           & \textbf{ 71.52\%}            &  \textbf{93.22\%}         & \textbf{80.94\%}           \\ 

\hline
\end{tabular}
\end{table}

Table \ref{tab:1} demonstrates that the ChatGPT-based models achieved the best performance across all metrics, significantly outperforming the baseline models such as BERT, RoBERTa, and XLNet. The ChatGPT-based model obtained the highest sensitivity, PPV, and F1-score, showcasing its superior capability to detect true positives and balance precision and recall. Notably, the performance of the two ChatGPT versions was remarkably similar, indicating the consistency and robustness of the ChatGPT-based approach for diagnosing social language disorders (SLDs). In contrast, the baseline models lagged behind in all metrics, highlighting the advantage of using advanced LLMs for this task.

\subsubsection{Evaluation of Speaker Diarization}

To further examine the impact of utilizing speaker diarization, we conducted ablation experiments that varied the use of speaker diarization tools integrated with our ChatGPT-based approach:

\begin{table}[h]
\centering
\caption{Ablation Experiments: Performance Metrics of Different Schemes}
\label{tab:2}
\begin{tabular}{|l|c|c|c|c|}
\hline
\textbf{Model} & \textbf{Accuracy} & \textbf{PPV} & \textbf{Sensitivity} & \textbf{F1 Score} \\ 
\hline
wo/ Speaker Diarization& 63.64\%           & 48.12\%            & 63.64\%         &   54.80\%           \\ 
w/ Pyannote& 68.18\%           & 59.68\%            & 68.18\%         & 60.45\%           \\ 
w/ Microsoft& 72.73\%           & 71.25\%            & 72.73\%         &  66.10\%           \\ 
w/ Human Diarization & \textbf{82.00\%}           & \textbf{91.06\%}            &  82.49\%         & \textbf{86.57\%} \\
\hline
\textbf{w/ Google(Ours)}            & 69.14\%           &  71.52\%            &  \textbf{93.22\%}         & 80.94\%          \\ 
\hline
\end{tabular}
\end{table}

Integrating Google’s speaker diarization (SD) technology with our ChatGPT-based model significantly improved all performance metrics, achieving the highest scores overall. Unlike other diarization tools, Google SD not only distinguishes multiple speakers (e.g., Speaker 1, Speaker 2) but crucially identifies whether the speaker is the examiner or the patient. This capability is especially valuable for tasks requiring an understanding of interaction dynamics and speaker roles (examiner vs. patient), which significantly influence the model’s performance in contextual analysis and response generation.

Together, these results validate the potential of incorporating sophisticated AI-driven tools like speaker diarization with ChatGPT to enhance the accuracy and efficiency of diagnostics.

\subsection{Analysis of Features of Language Deficit}
\subsubsection{Correlation Analysis}

\begin{figure}
 \centering
 \includegraphics[width=0.8\linewidth]{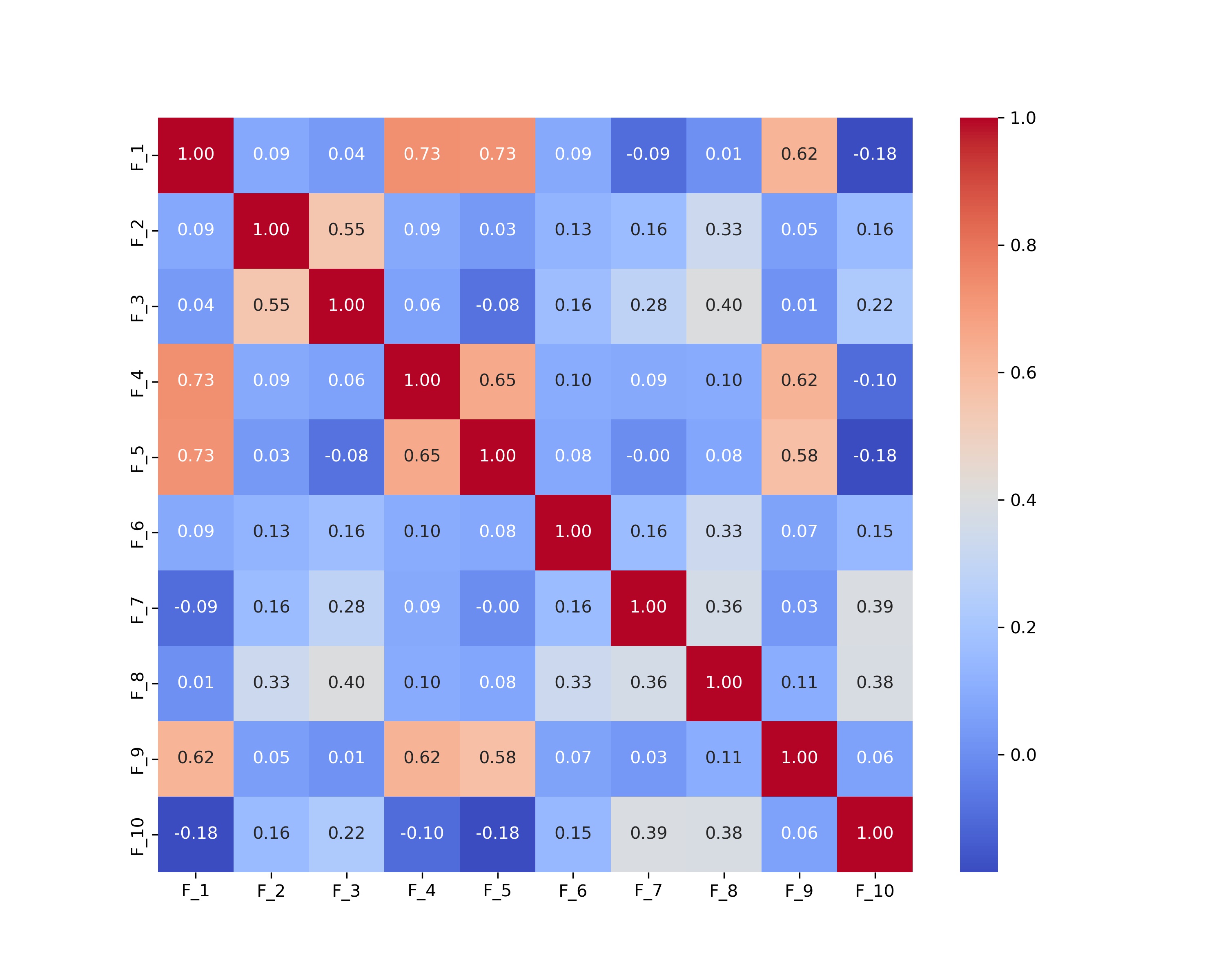}
 \caption{Correlation Coefficients Between Features of Language Deficit}
  \label{fig:0}
\end{figure}

To thoroughly understand the interrelationships between different language features identified in the ASD diagnostic assessments, We analyzed the interrelationships among ten language features ($F_{1}$ to $F_{10}$) derived from the Caltech dataset. These features represent various aspects of language use that may indicate ASD, such as repetitive use of words or unusual language patterns. We calculated Pearson correlation coefficients between each pair of features to quantify their linear relationships. Each language feature is represented as a binary variable, where '1' indicates the presence and '0' indicates the absence of that specific language disorder feature within any given sample. For example, if a feature detected by ChatGPT such as "echolalic repetition" is observed in the dialogue during a diagnostic session, it is marked as '1' for that session; otherwise, it is marked as '0'. This binary coding allows us to apply Pearson correlation to measure the linear relationship between each pair of features across all samples. This analysis pinpoints which features often co-occur within the linguistic profiles of ASD diagnosed through ADOS-2, Module 4. The computed correlation matrix for the features is presented in Figure \ref{fig:0}.

\begin{enumerate}
  \item \textbf{Highly Correlated Features:}
  \begin{itemize}
    \item \(F_1\), \(F_4\), and \(F_5\): These features show very high correlations ($r = 0.734$ between $F_1$ and $F_4$, $r = 0.727$ between $F_1$ and $F_5$, and $r = 0.655$ between $F_4$ and $F_5$). This suggests they may capture similar aspects of linguistic behavior, possibly related to the repetitive or stereotyped use of language, which is a common indicator of ASD.
    \item \(F_4\) and \(F_9\): Another pair, $F_4$ and $F_9$ ($r = 0.622$), indicates a strong association, which might reflect overlapping features of language presentation in ASD, such as idiosyncratic language use or atypical language processing.
  \end{itemize}

  \item \textbf{Moderately Correlated Features:}
  \begin{itemize}

    \item \(F_2\) and \(F_3\): These features exhibit moderate correlations (r = 0.55 for F2 and F3), with F2 and F3 showing a correlation strength that is typically described as 'moderate' according to established guidelines for interpreting correlation coefficients. These correlations suggest a meaningful but not strong relationship in linguistic traits, such as variability in speech that includes both repetitive and novel elements.    
  \end{itemize}

  \item \textbf{Negatively Correlated Features:}
  \begin{itemize}
    \item \(F_1\) and \(F_{10}\): The negative correlation ($r = -0.184$) suggests that when $F_1$ (possibly denoting less severe ASD indicators) is present, $F_{10}$ (perhaps denoting more severe ASD indicators) is less likely to be present, and vice versa. This can help differentiate levels of language impairment in ASD diagnoses.
  \end{itemize}
\end{enumerate}

Together, the exploration of the correlation between features of language deficit in ASD offers valuable insights into the complex nature of communication challenges faced by individuals on the spectrum.

\subsubsection{Distribution of Features of Language Deficit  Across Scenarios}

\begin{table}[h]
\centering
\caption{Prevalence of Linguistic Features by Scenario Indicative of Language Deficits in ASD}
\begin{tabular}{lcccccccccc}
\toprule
Scenario & F1    & F2    & F3    & F4    & F5    & F6    & F7    & F8    & F9    & F10   \\
\midrule
3  & 0.45 & 0.64 & 0.52 & 0.32 & 0.39 & 0.59 & 0.48 & 0.41 & 0.39 & 0.36 \\
4  & 0.57 & 0.59 & 0.45 & 0.32 & 0.41 & 0.59 & 0.57 & 0.39 & 0.30 & 0.32 \\
5  & 0.43 & 0.55 & 0.41 & 0.23 & 0.43 & 0.61 & 0.48 & 0.27 & 0.25 & 0.32 \\
6  & 0.41 & 0.34 & 0.25 & 0.18 & 0.36 & 0.55 & 0.43 & 0.23 & 0.07 & 0.30 \\
7  & 0.48 & 0.39 & 0.34 & 0.20 & 0.36 & 0.50 & 0.39 & 0.30 & 0.20 & 0.30 \\
9  & 0.57 & 0.43 & 0.30 & 0.20 & 0.32 & 0.48 & 0.41 & 0.23 & 0.16 & 0.20 \\
11 & 0.43 & 0.39 & 0.32 & 0.14 & 0.32 & 0.59 & 0.55 & 0.32 & 0.18 & 0.32 \\
12 & 0.45 & 0.55 & 0.41 & 0.18 & 0.36 & 0.55 & 0.61 & 0.41 & 0.07 & 0.32 \\
13 & 0.55 & 0.36 & 0.25 & 0.02 & 0.36 & 0.48 & 0.30 & 0.25 & 0.02 & 0.23 \\
14 & 0.36 & 0.41 & 0.18 & 0.11 & 0.23 & 0.34 & 0.20 & 0.11 & 0.07 & 0.11 \\
15 & 0.64 & 0.57 & 0.32 & 0.36 & 0.52 & 0.64 & 0.57 & 0.32 & 0.30 & 0.20 \\
\bottomrule
\end{tabular}
\label{tab:feature_distribution}
\end{table}

This subsection analyzes the correlations between linguistic features across various ADOS scenarios to identify patterns that may indicate language disorders associated with ASD. The focus is on scenarios with direct dialogue between the examiner and the patient, reflecting our study’s emphasis on communication. While Scenarios \(S_1\), \(S_2\), \(S_8\), and \(S_{10}\) provide valuable insights into various aspects of cognitive and social functioning, they were not included in this analysis due to their lack of direct dialogue-based interaction between the examiner and the patient, which is a primary focus of our research. To effectively analyze the differences in the distribution of values from $F_{1}$ to $F_{10}$ across various scenarios, we conducted a detailed statistical examination (see Table \ref{tab:feature_distribution}). This analysis helps to understand how the prevalence of each linguistic feature associated with ASD varies across the scenarios, which can provide insights into the contexts or conditions under which certain features are more likely to appear.

    

We have derived the following observations and insights:
\begin{itemize}
    \item \textbf{Feature Prevalence:} The occurrence rates of features \(F_2\), \(F_6\), and \(F_7\), which represent aspects of unconventional content, verbal fluency, and excessive social phrasing respectively, were consistently above 60\% across most scenarios. This high prevalence underscores their significance as key indicators of ASD. 

    \item \textbf{Language Feature in Social Contexts:} Features such as \(F_1\) (possibly related to echolalia or repetitive speech), \(F_4\), \(F_5\), and \(F_9\) (potentially related to atypical or stereotyped language use) were entirely absent in several scenarios, underscoring their sensitivity to specific social or communicative contexts.

    \item \textbf{Scenario-Specific Patterns:} High prevalence rates in \(F_7\) during the \(S_7\) (i.e., 'Emotions') scenario, and diverse responses in \(F_2\) across the \(S_{12}\) (i.e., 'Friends, Relationships, and Marriage')   and  \(S_{15}\) (i.e., 'Creating a Story')  scenarios suggest that certain linguistic features were particularly elicited by emotional or social relational contexts.
\end{itemize}

To further demonstrate the utility of this analysis, we specifically focused on the \(S_3\) and \(S_9\) scenarios, which are essential for evaluating narrative skills and abstract reasoning, respectively. We derived the following results:

\begin{figure}[h]
 \centering
 \includegraphics[width=0.8\linewidth]{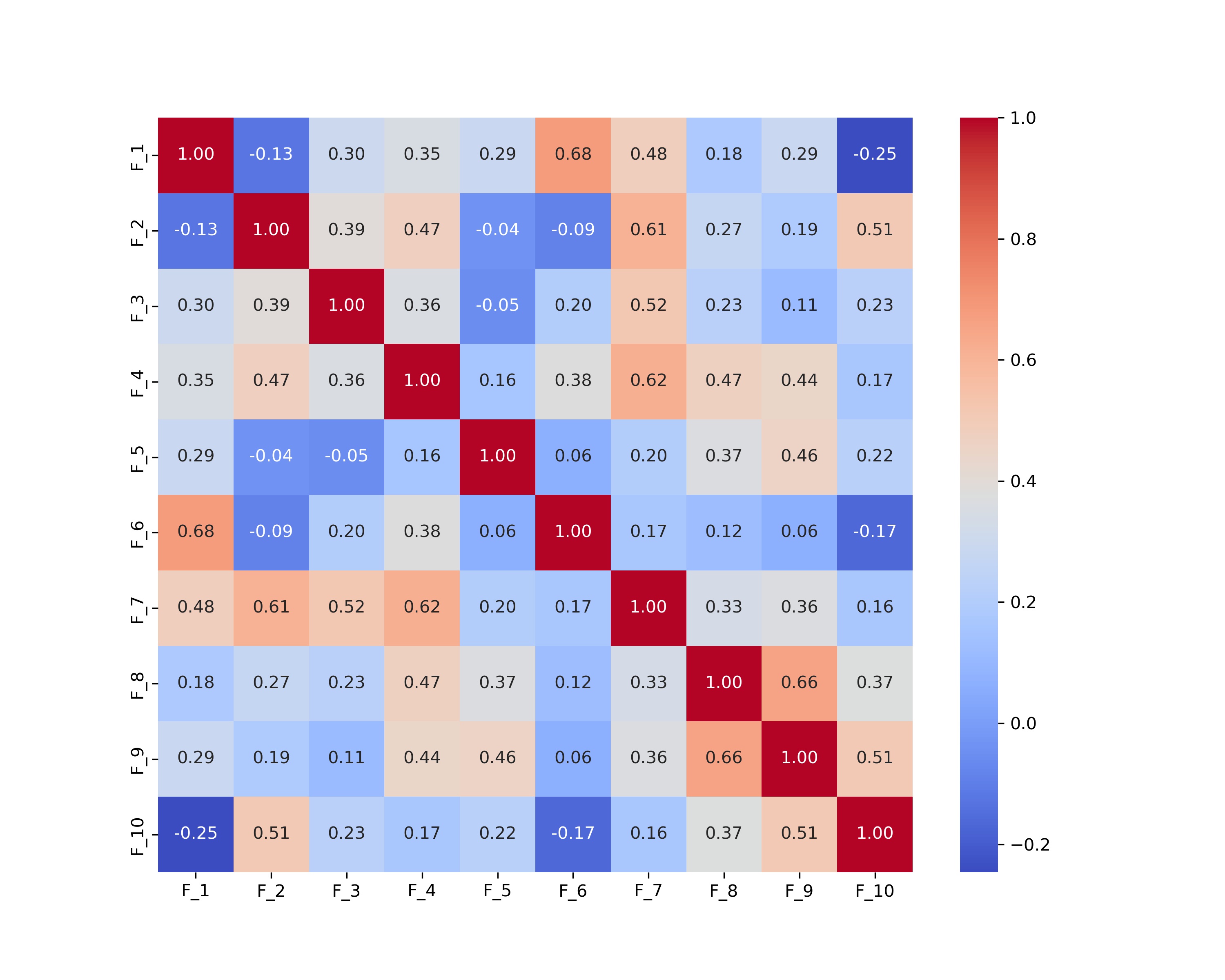}
 \caption{Correlation matrix of linguistic features \(F_{1}\) to \(F_{10}\) in the \(S_3\) (i.e., 'Description of a Picture') scenario. The matrix shows strong correlations between features, underscoring the interdependencies that influence how visual information is described.}
  \label{fig:2}
\end{figure}

\begin{figure}[t]
 \centering
 \includegraphics[width=0.8\linewidth]{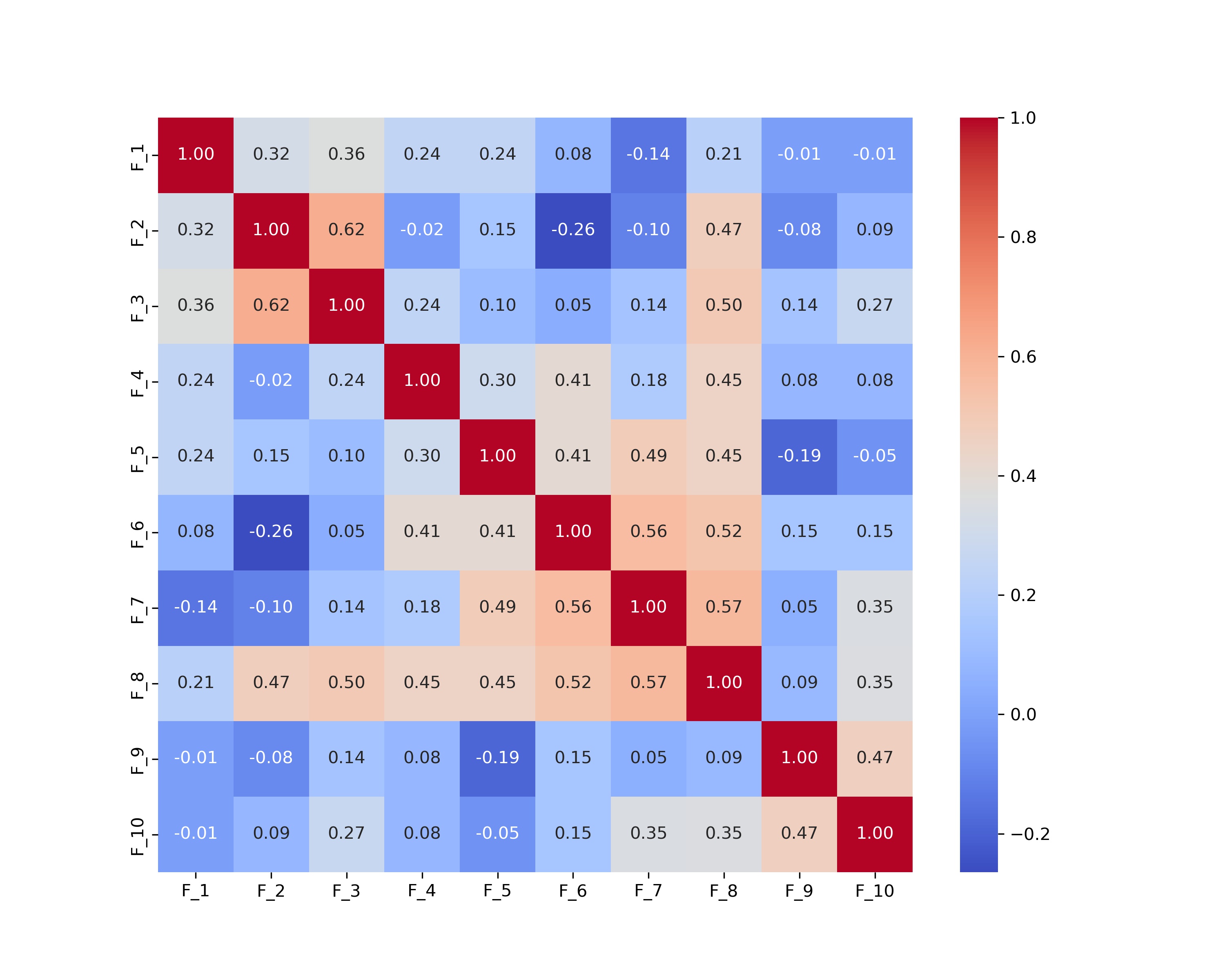}
 \caption{Correlation matrix of linguistic features \(F_{1}\) to \(F_{10}\) in the \(S_9\) (i.e., 'Cartoons') scenario. This matrix highlights correlations that elucidate the cognitive and perceptual challenges in interpreting cartoons, essential for understanding narrative contexts and humor in ASD.}
  \label{fig:3}
\end{figure}

\begin{itemize}
    \item \textbf{Scenario \(S_3\)} (Figure \ref{fig:2}): The strong correlation between \(F_1\) and \(F_6\) (0.68) indicates challenges in effectively summarizing visual content. This may reflect difficulties in processing and conveying information succinctly, which is often a challenge for individuals with ASD. A high correlations (0.66) between \(F_8\) (monotone social expression) and \(F_9\) (stereotyped media quoting) suggests that individuals may struggle with varying their emotional expressions, which could affect the emotional richness of their speech. Correlations between \(F_7\)  (excessive social phrasing) and \(F_4\) (incongruous humor) (0.62), and between \(F_7\) and \(F_2\) unconventional content (0.61) suggest a connection between repetitive social expressions and the production of either inappropriate humor or atypical content. This pattern may indicate that individuals with ASD use scripted language as a strategy to manage social interactions, although this can often result in conversations that seem awkward or misplaced.
    
    \item \textbf{Scenario \(S_9\)} (Figure \ref{fig:3}): The strong correlation between \(F_2\) and \(F_3\) (0.62) suggest that individuals with ASD might struggle to adjust their language to fit the context appropriately. This is particularly problematic in scenarios like watching and discussing cartoons, where understanding shifting dialogues and multiple characters' perspectives is essential; Additionally, \(F_8\)’s significant correlations with \(F_7\) (0.57) reflects difficulties in varying emotional tone and using phrases that might be socially appropriate. Individuals exhibiting these features tend to speak in a flat, unmodulated manner while possibly overusing certain social phrases, making their speech seem rigid and scripted. Such speech patterns can make it difficult for them to engage in spontaneous and emotionally responsive interactions, which are critical for successful social exchanges. 
    
\end{itemize}

Together, these correlations suggest co-morbid linguistic challenges that individuals with ASD may encounter in scenarios requiring detailed visual interpretation or complex narrative understanding. 

\subsubsection{Feature Detection Comparison Across Models}

\begin{figure}[h]
 \centering
 \includegraphics[width=0.6\linewidth]{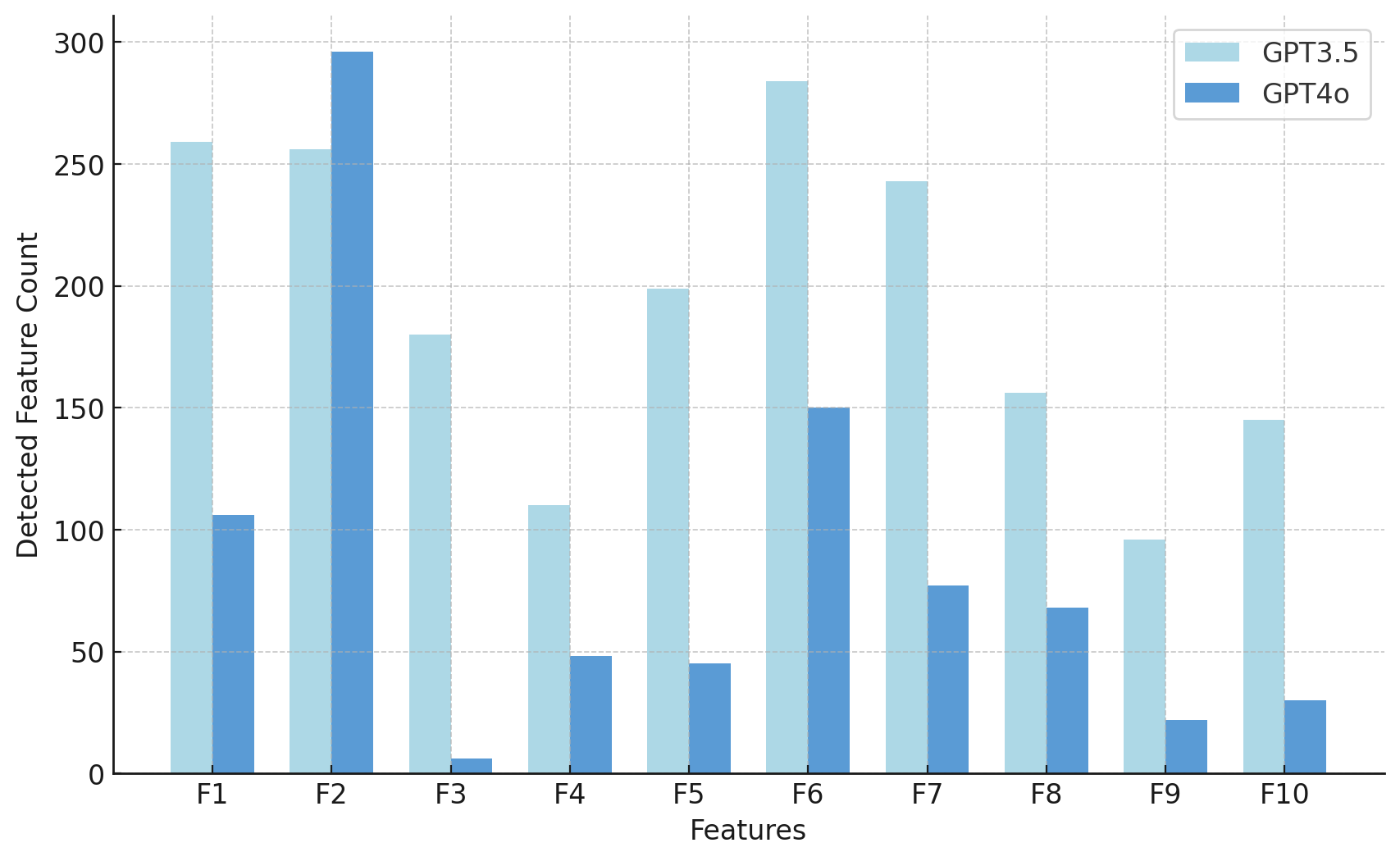}
 \caption{Detected SLD Feature Count Comparison between GPT-3.5 and GPT-4o}
  \label{fig:00}
\end{figure}

To investigate how different features influence model performance, we further analyze the performance of the models through a feature detection comparison. Figure \ref{fig:00} provides a histogram comparing the counts of specific linguistic features related to SLDs detected by GPT3.5-based and GPT4o-based models.
 
GPT3.5 shows higher detection rates across key features like F2 (unconventional content), F6 (verbal fluency), and F7 (excessive social phrasing), which are crucial for diagnosing language deficits associated with ASD. GPT4o performs well in identifying features such as F3 (pronoun displacement) and F10 (clichéd verbal substitutions), but overall detects fewer features compared to GPT3.5. This histogram highlights GPT3.5’s sensitivity to a broader range of linguistic nuances, supporting its superior accuracy, PPV, and sensitivity. By detecting critical indicators of SLDs, GPT3.5 offers a more comprehensive diagnostic tool for analyzing ASD-related communication challenges.

\subsection{Case Study}

Lastly, we present two case studies to illustrate the practical application of ChatGPT in identifying language deficits in ASD.

Table \ref{tab:5} and \ref{tab:6} show the dialogue between an examiner (E) and a patient (P). It showcases typical conversational challenges faced by individuals with ASD. The patient's responses highlight several linguistic features that indicate underlying language disorders.

\begin{table}[h]
\centering
\caption{Case Study Analysis: Identifying Language Deficits in an Examiner-Patient Dialogue. Phrases highlighted in \textcolor{blue}{blue} indicate observed linguistic anomalies, while \textcolor{red}{red} underscores the specific feature category of language deficits.}
\label{tab:5}
\begin{tabular}{p{0.95\textwidth}}
\hline
\textbf{Examiner (E) - Patient(P) Dialogue} \\
\hline
E: Okay. So, do you have some friends? \\ 
\textbf{P: Uh, $\textcolor{blue}{\text{do I have some friends?}}_{\textcolor{red}{F_{1}}}$} \\
E: Um-hum. \\
\textbf{P: $\textcolor{blue}{\text{Well, pretty much}}_{\textcolor{red}{F_{10}}}$, three of them from peers.} \\
E: Three of them from here?\\
\textbf{P: Um-hum.} \\
E: Okay. Can you tell me a little bit about them? \\
\textbf{P: Well, they're kind of living near, they kind of live near $\textcolor{blue}{\text{her}}_{\textcolor{red}{F_{3}}}$ farther from here.} \\
E: They're further from here? What do they like? \\
\textbf{P: $\textcolor{blue}{\text{What do they like?}}_{\textcolor{red}{F_{1}}}$ They're kind of energetic, just like me. Cool.}\\
E: Um, and what do you guys like to do together? Man, we like movies and stuff.\\
\textbf{P: And you've got to know them through peers.} \\
...\\
E: And, but you said you'd go, you'd like to go to movies and stuff as well. Do you go to movies with them, or? \\
\textbf{P: We use $\textcolor{blue}{\text{gas movies}}_{\textcolor{red}{F_{9}}}$. }\\
E: Oh, you talk about it? \\
\textbf{P: Yeah.} \\
E: Okay. And are there people outside of peers that you're friends with, or? \\
\textbf{P: You mean, uh, $\textcolor{blue}{\text{outside of peers}}_{\textcolor{red}{F_{1}}}$?} \\
...\\
\textbf{P: Crazy, crowded, crooked.. One of those years people.}\\
E: Oh. \\
\textbf{P: They $\textcolor{blue}{\text{triggered the trip, pregnant sound effects.}}_{\textcolor{red}{F_{9}}}$} \\
E: Oh, yeah. \\
\textbf{P: When they asked about dating.. Um What, where do you, uh, want to live when you get older }\\
E: face? \\
\textbf{P: I want to live in $\textcolor{blue}{\text{ a lounge and dirty autistic matching }}_{\textcolor{red}{F_{2}}}$. You know, you can $\textcolor{blue}{\text{ zillion blocks matching}}_{\textcolor{red}{F_{10}}}$.} \\
E: And who do you think you would like to live with, with your family or roommates or by yourself? \\
\textbf{P: I want to live with my family there. Okay} \\
... \\
\hline
\textbf{Extracted Features Based on ChatGPT Response} \\
\hline
\hspace{1em} \textbf{Echoic Repetition ($F_{1}$)}: When the examiner first asked the patient whether they have some friends, the patient echoed the question back at the examiner before answering. In subsequent interactions, the patient frequently mimics the examiner's questions verbatim before answering. \\
\hline
\hspace{1em} \textbf{Unconventional Content ($F_{2}$)}: The patient refers to living in "a lounge and dirty autistic matching" rather than using any conventional descriptions for living spaces. Similarly, the phrase "zillion blocks matching" has an unusual content. \\
\hline
\hspace{1em} \textbf{Pronoun Displacement ($F_{3}$)}: The patient referred to his own house as "her". \\
\hline
\hspace{1em} \textbf{Stereotyped Media Quoting ($F_{9}$)}: The patient quoted "gas movies" and "triggered the trip, pregnant sound effects", which seems to be quoted from an external media source. \\
\hline
\hspace{1em} \textbf{Clichéd Verbal Substitutions ($F_{10}$)}: The patient uses clichéd expressions like "well, pretty much" instead of giving direct responses. \\
\hline
\end{tabular}
\end{table}

\begin{table}[h]
\centering
\caption{Case Study Analysis: Identifying Language Deficits in an Examiner-Patient Dialogue. Phrases highlighted in \textcolor{blue}{blue} indicate observed linguistic anomalies, while \textcolor{red}{red} underscores the specific feature category of language deficits.}
\label{tab:6}
\begin{tabular}{p{0.95\textwidth}}
\hline
\textbf{Examiner (E) - Patient(P) Dialogue} \\
\hline
E: So, I'm going to ask you a few questions about work and school \\ 
\textbf{P: Yes.} \\
E: Um, first of all, do you have a job? \\
\textbf{P: No, I used to be laid off.} \\
... \\
E: And that's okay? Yeah Um, while you were working or now at school, or at high school, maybe before that, did you have a group of, any problems getting along with people You weren't in high school? \\
\textbf{P: Any school. Well, like, like, 
$\textcolor{blue}{\text{stupid schools for you when I was developing angry or high school.}}_{\textcolor{red}{F_{2}}}$} \\
...\\
E: What kind of things you used to bother you that other people did? \\
\textbf{P: Like, uh, when I was in the school bus I had students grabbing my backpack, whatever, and $\textcolor{blue}{\text{I didn't mad it or suck.}}_{\textcolor{red}{F_{10}}}$}   \\
...\\
E: And have you ever done anything so that other people wouldn't teach soon?  \\
\textbf{P: Yes, but sometimes they just, it's like they've been doing it for a while, so it's just kind of like Hey, you know $\textcolor{blue}{\text{or what, whatever,}}_{\textcolor{red}{F_{6}}}$ we'll just tease him about something else. } \\
... \\
\hline
\textbf{Extracted Features Based on ChatGPT Response} \\
\hline
\hspace{1em} \textbf{Unconventional Content ($F_{2}$)}: There are instances where the patient uses unconventionally chosen phrases like "stupid schools for you when I was developing angry or high school". \\
\hline
\hspace{1em} \textbf{Superfluous Phrase Attachment ($F_{6}$)}: The patient attaches redundant phrases or filler expressions to their speech without contributing any substantive meaning or context, such as 'whatever' and 'or whatever'. \\
\hline
\hspace{1em} \textbf{Clichéd Verbal Substitutions ($F_{10}$)}: The patient resorts to clichéd expressions when describing how he felt during certain situations: "I didn't mad it or suck. \\
\hline

\end{tabular}
\end{table}

Together, these case studies demonstrate the effectiveness of using ChatGPT, combined with structured conversational analysis, to diagnose social language disorders in ASD. The identified features align well with known ASD communication challenges, providing a robust basis for further diagnostic evaluation and intervention planning.

\section{Discussion}
\subsection{Leveraging ChatGPT for ASD Diagnosis}
The experimental results demonstrate ChatGPT’s exceptional performance in diagnosing ASD-related language deficits, particularly through significant improvements in sensitivity and PPV. High sensitivity reflects the model’s ability to identify a substantial proportion of true positives, ensuring that fewer ASD cases are missed. This is critical in clinical settings, where delayed diagnosis can hinder timely interventions and exacerbate social and communicative challenges in patients. ChatGPT’s ability to capture subtle linguistic anomalies, such as repetitive phrasing and pronoun reversal, further showcases its capacity to generalize across diverse linguistic patterns while identifying ASD-specific markers. Similarly, the model’s high PPV underscores its reliability by reducing false positives. This precision is particularly valuable in distinguishing ASD-related language deficits from typical developmental variations, which often present overlapping characteristics. By effectively minimizing misdiagnoses, ChatGPT alleviates undue stress for patients and ensures optimal allocation of clinical resources.

These results highlight ChatGPT’s potential to revolutionize ASD diagnostics in medical practice. Its high sensitivity ensures that at-risk individuals are identified early, paving the way for timely therapeutic interventions that can improve outcomes significantly. Meanwhile, its elevated precision supports confident decision-making, enabling clinicians to focus on cases most likely to benefit from tailored care. Additionally, the interpretable outputs generated by ChatGPT provide actionable insights into linguistic features, such as echolalia and atypical language use, offering a valuable tool for designing personalized treatment plans. The ability to integrate AI-driven diagnostics into existing workflows not only streamlines the diagnostic process but also sets a foundation for more consistent and scalable practices in developmental medicine.

\subsection{Impact of Speaker Diarization on ASD Diagnosis}
The integration of speaker diarization, particularly Google’s role-based system, has been pivotal in enabling automatic diagnosis of ASD in conversational scenarios. By distinguishing examiner and patient speech, the model is better equipped to analyze patient-specific language patterns. This role differentiation significantly contributes to the model’s high sensitivity and contextual understanding.

However, our results indicate that the performance of automatic speaker diarization, while effective, still falls short compared to human diarization. In ablation experiments, human diarization achieved the highest Accuracy (82.00\%), PPV (91.06\%), and  F1 score (86.57\%), outperforming automated methods. This highlights the critical role of speaker diarization in ASD diagnosis and suggests that further advancements in diarization technology are likely to improve diagnostic accuracy. As automated diarization tools like Google’s system continue to evolve, their ability to match or exceed human-level precision will make them even more valuable for diagnosing ASD, where conversational context and speaker roles are essential.

\subsection{Scenario-Based Analysis of Social Language Disorder Features}

The prevalence of language deficits across different scenarios revealed valuable patterns about how individuals with ASD respond to various social and cognitive contexts:

\begin{itemize}
    \item Emotional and Creative Contexts: Scenarios such as $S_{7}$ (Emotions) and $S_{15}$ (Creating a Story) elicited higher occurrences of features like $F6$ (verbal fluency) and $F_{7}$ (excessive social phrasing). This suggests that emotional and narrative-based interactions highlight specific communication challenges in individuals with ASD, such as difficulty adapting language to dynamic social contexts or producing coherent narratives \cite{losh2014quantifying}.
    
    \item Visual Interpretation Challenges: In $S_{3}$ (Description of a Picture), strong correlations between $F_{1}$ (repetition) and $F_{6}$ (verbal fluency) indicate struggles with organizing and summarizing visual information. These patterns align with known cognitive challenges in ASD, such as difficulties in abstraction and the integration of sensory input into coherent verbal descriptions \cite{happe2006weak}.
    
    \item Abstract Reasoning in Narratives: The results from $S_{9}$ (Cartoons) highlight co-occurring features like $F_{2}$ (unconventional content) and $F_{3}$ (pronoun displacement), suggesting that interpreting multi-character narratives and shifting perspectives poses significant difficulties for individuals with ASD. These findings reinforce the need for scenario-specific evaluations that consider how different contexts elicit unique linguistic deficits.
    
\end{itemize}

\subsection{Clinical Implications for ASD Diagnosis}
The results from the present study have strong clinical implications. Our model's ability to analyze complex language interactions is valuable in clinical settings. It allows for an earlier and more accurate detection of language deficits, which are often indicative of ASD. This early diagnosis is crucial for the timely intervention that can lead to better management and outcomes for adults with ASD. A promising future direction is to leverage the power of LLM into distinguishing ASD from other language impairment during the development \cite{loucas2008autistic}. If LLM can shed new insight to the co-occurrence of ASD and language impairment, clinical diagnosis of developmental disorders might benefit from human-AI collaboration.

\subsection{Limitation and Future Work}
Despite its innovations, this study has limitations that should be addressed in future research. One limitation is the relatively small dataset size, while sufficient for exploratory analyses, restricts the statistical power and generalizability of our findings. As medical data is often difficult to obtain due to privacy and accessibility constraints, future work will prioritize acquiring larger and more diverse datasets to enhance the robustness and applicability of our approach. Additionally, the effectiveness of the ChatGPT model depends heavily on the quality and variety of the training data. The current dataset represents a relatively homogeneous population in terms of linguistic and cultural backgrounds. Expanding the dataset to include a more diverse population could improve the model's robustness and generalizability. Furthermore, future studies could explore the integration of multimodal data analysis \cite{han2022multimodal} to enhance diagnostic capabilities further. Combining speech with visual cues such as facial expressions and body language could provide a more comprehensive view of an individual's communicative and social behaviors. Finally, refining the models to incorporate feedback loops that enable continual learning from new data could adaptively improve diagnostic accuracy over time \cite{biesialska2020continual}. These enhancements will pave the way for more robust and comprehensive diagnostic tools in the future.

\section{Methods}

\subsection{Framework for Diagnosing Autism and Identifying Language Disorders}

Building on the foundational practices established by the ADOS-2, specifically Module 4 designed for verbally fluent adolescents and adults, we have developed a comprehensive framework (see Figure \ref{fig:0}) that incorporates LLMs like ChatGPT. This framework is tailored to enhance the diagnostic PPV and identification of language disorders in individuals suspected of having ASD. Specifically, it involves the following components:

\begin{figure}[h]
 \centering
 \includegraphics[width=1\linewidth]{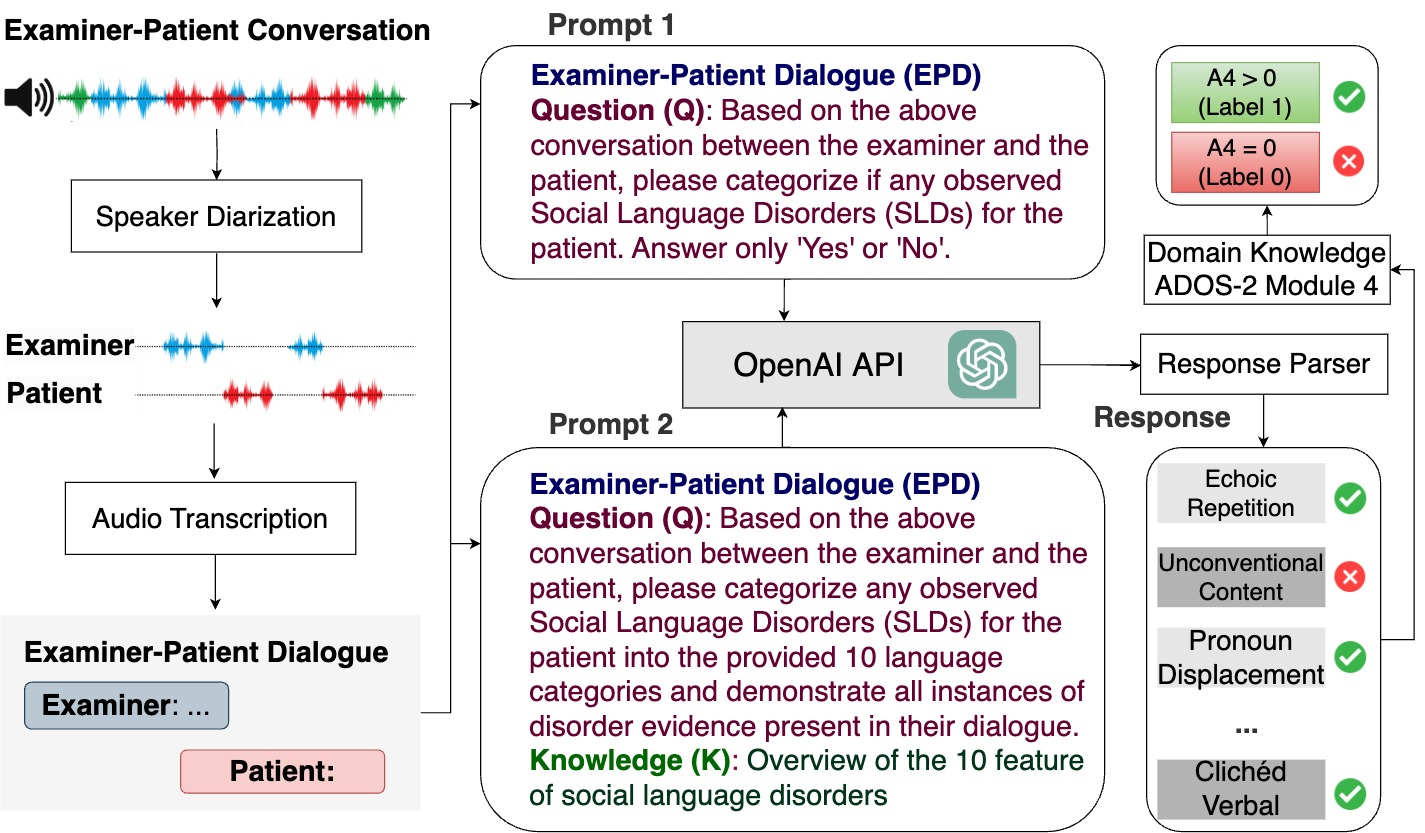}
 \caption{Framework for Diagnosing Autism and Identifying Language Disorders. A4 Score: Stereotyped/Idiosyncratic Use of Words or Phrases (see Table \ref{table:A4Score}).}
  \label{fig:4}
\end{figure}

\begin{itemize}
    \item \textbf{Speaker Diarization and Audio Transcription.}
This technology segments the audio recordings to precisely separate the speech of the examiner from that of the patient. Such separation is crucial as it enhances the understanding of the patient's behavior in conversational contexts by isolating their verbal responses, which are then analyzed for potential linguistic abnormalities. The audio segments identified through diarization are subsequently transcribed into text using Google's state-of-the-art transcription technologies (see Fig. \ref{fig:4}). This conversion facilitates a detailed examination of the social language used by the patient, aiding in the detection of disorder-specific features within their speech.

\item \textbf{Language Pattern Analysis Using ChatGPT.}

In this framework, ChatGPT is utilized not only to diagnose ASD but also to identify specific language disorder characteristics that are indicative of ASD. Structured prompts are prepared to gather key information from examiner-patient dialogues. These prompts are crafted as follows:

\begin{itemize}
    \item \textbf{Examiner-Patient Dialogue (EPD):} The dialogue text, which includes conversational exchanges between the examiner and the patient, serves as the primary input for ChatGPT. This dialogue is carefully processed to maintain the integrity and context of the interaction, ensuring that all relevant linguistic cues are preserved.

    \item \textbf{Question Design:} To guide the analysis, specific questions are formulated based on the dialogue content. These questions aim to direct ChatGPT’s attention to potential signs of language disorders, such as repetitive phrasing, atypical language use, or disrupted conversational flow.

    \item \textbf{Knowledge Design:} This component incorporates domain-specific knowledge from autism diagnostics, which is used to refine ChatGPT's responses. By integrating expert knowledge, the model is better equipped to recognize and interpret the subtle nuances that characterize ASD-related language disorders.

    \item \textbf{Prompt Integration:} The complete prompt for ChatGPT includes the dialogue text, the targeted questions, and the expert knowledge cues. This integrated approach helps in precisely pinpointing disorder characteristics that might be overlooked in a less structured analysis.
\end{itemize}

\item  \textbf{Functionality of the Response Parser:}
The response parser is a critical element that processes the outputs from ChatGPT. It performs two main functions:

\begin{itemize}

    \item \textbf{Identification of Specific Characteristics:} Beyond mere diagnosis, the response parser identifies specific characteristics of the language disorders. It extracts detailed information about the nature and extent of the linguistic anomalies detected, such as the type of stereotypy or idiosyncrasy in the patient's speech.
\end{itemize}

    \item \textbf{Diagnosis based on ADOS-2 Module 4:} The final classification module evaluates the detected features (F1-F10) based on predefined criteria derived from ADOS-2 guidelines. This module integrates domain knowledge to predict A4 scores (A4 = 0 or A4 $>$ 0) by systematically analyzing the presence and severity of SLDs as follows:
    \begin{itemize}
        \item If critical features, such as F1 (Echoic Repetition) or F9 (Stereotyped Media Quoting), are identified, the dialogue is classified as A4 $>$ 0 (Label 1). These severe features strongly indicate significant social language impairment.

        \item Cumulative Features Rule: If more than two other related features (e.g., F2, F3, etc.) are present, the module predicts A4 $>$ 0 (Label 1). This reflects the cumulative impact of multiple moderate symptoms.

        \item Non-SLD Classification: If neither severe symptoms nor a sufficient number of related features are detected, the module predicts A4 = 0 (Label 0). This outcome suggests no notable social language disorder in the dialogue.

        \item A4 Classification Output: Based on the above rules, the module outputs the A4 classification results: A4 = 0 (Label 0): No significant social language disorders detected. A4 $>$ 0 (Label 1): Significant social language disorders detected.
   \end{itemize}

\end{itemize}

\subsection{Speaker Diarization and Transcription for Examiner-Patient Interactions}

Accurate separation and transcription of examiner-patient dialogues are essential for analyzing linguistic patterns in ASD diagnosis. In this study, we leverage speaker diarization and transcription technologies to automate this process, addressing the limitations of traditional manual annotation. Speaker diarization involves segmenting audio recordings into distinct sections attributed to individual speakers, ensuring precise identification of patient responses while maintaining the conversational context provided by the examiner.

We employed Google's advanced role-based diarization system, which extends beyond generic speaker labeling (e.g., "Speaker 1" or "Speaker 2") by explicitly identifying roles as "Examiner" and "Patient." This functionality is particularly valuable in ASD diagnostics, as it enables the focused analysis of patient-specific language behaviors. Following diarization, the segmented audio files were transcribed into text, creating a structured dataset for linguistic feature extraction and SLD analysis. The workflow includes three key steps:

\begin{itemize}
    \item Audio Segmentation: Long recordings were divided into segments based on diagnostic scenarios, such as picture description or abstract reasoning tasks. This segmentation ensures targeted analysis for each task.
    
    \item Speaker Role Labeling: Using Google's Medical-Conversation Model, the diarization system identified and labeled speaker roles within the conversation, preserving the dynamic interplay between examiner prompts and patient responses.
    
    \item Speech-to-Text Transcription: Each audio segment was transcribed into text with speaker-specific annotations, forming a reliable foundation for subsequent computational analysis.
    
\end{itemize}

This automated approach not only reduces the manual workload but also ensures consistent and scalable processing of examiner-patient dialogues, laying the groundwork for robust ASD diagnostic models. By integrating diarization and transcription into a unified pipeline, we streamline the analysis of conversational interactions while maintaining alignment with clinical diagnostic frameworks like ADOS-2.

\subsection{Diagnosing Language Disorders Associated with Autism via ChatGPT}

This subsection details the methodology used to employ ChatGPT, an advanced language model, for diagnosing SLDs in individuals with ASD. The approach leverages a structured prompt to analyze dialogues between examiners and patients, determining the presence of communicative impairments characteristic of ASD.

\textbf{Prompt Design for ChatGPT: } This process involved parts. In Part 1, Examiner-Patient Dialogue (EPD), the input to ChatGPT included the transcribed dialogue between the examiner and the patient, presenting the conversational context needed for assessment. In Part 2, Question (Q), following the dialogue, ChatGPT was asked: "Based on the above conversation between the examiner and the patient, please categorize if any observed SLDs for the patient. Answer only 'Yes' or 'No'." This question aimed to elicit a definitive response based on the dialogue's content, focusing solely on the presence or absence of disorder indicators.

\textbf{Response Interpretation Using ChatGPT: }
The responses from ChatGPT were parsed to determine the presence of SLDs. The decision process was as follows:

\begin{itemize}
    \item \textbf{Response Parser:} Each response from ChatGPT, indicating either affirmation ("Yes") or negation ("No"), was analyzed to ascertain whether the patient exhibited symptoms of SLDs based on the dialogue provided. The parser specifically looked for expressions of affirmation or negation concerning the presence of communicative impairments.
    
    \item \textbf{Diagnosis Determination:} For each subject, a diagnosis of a SLD was considered positive if there was at least one scenario where ChatGPT affirmed the presence of SLDs ("Yes"). Conversely, if all scenarios resulted in a "No" from ChatGPT, the subject was not considered to have SLDs as per the dialogues analyzed.
\end{itemize}

By integrating ChatGPT's advanced analytical capabilities, this methodology refines the diagnostic process for social language disorders in ASD, enhancing both the efficiency and accuracy of assessments. This approach not only supports clinicians by providing a reliable diagnostic tool but also contributes to the broader field of psycholinguistics by offering insights into the communicative impairments often seen in ASD.

\subsection{Identifying Language Disorder Features Associated with ASD via ChatGPT}

This subsection elaborates on the methodology employed to harness ChatGPT for identifying specific language disorder features associated with ASD, guided by expert knowledge integrated from the ADOS-2, Module 4. The approach utilizes a comprehensive list of language disorders designed around the nuanced communication requirements and symptoms observed in verbally fluent adolescents and adults.


To facilitate the extraction of these  features of language deficits using ChatGPT, a specific prompt structure is utilized, as shown in 'Prompt 2' (Figure \ref{fig:4}). The prompt was organized into three parts to optimize the analysis, including Part 1: Examiner-Patient Dialogue (EPD), Part 2: Question (Q) - ..., and Part 3: Knowledge (K) - "Overview of the 10 features of social language disorders identified by ADOS-2 examiners, as shown in the column 'Explanation' in Table \ref{tab:3}.

This structured prompt design guides ChatGPT to analyze the transcribed conversations and categorize language deficits, enhancing the PPV of diagnostics based on observed linguistic patterns.

\textbf{Response Interpretation Using ChatGPT:}
ChatGPT's responses were analyzed to determine the presence and types of SLD features as follows:

\begin{itemize}
    \item \textbf{Response Parser:} The parser reviewed ChatGPT’s responses, which involved multiple labels corresponding to the 10 predefined SLD features. Each piece of dialogue could yield several labels, reflecting the multi-dimensional nature of language disorders.
    
    \item \textbf{Feature Classification:} Each response was predicted into multiple categories, constituting a multi-label classification task. This approach allowed for a comprehensive assessment of the patient's language abilities, identifying multiple SLD features from a single excerpt of dialogue.
\end{itemize}

The multi-label classification approach offers significant advantages in the diagnosis of SLDs. By categorizing dialogue into multiple SLD-related features, ChatGPT enables a comprehensive analysis of the patient’s communicative impairments, providing a nuanced understanding of their specific challenges. This detailed insight is critical for ensuring accurate diagnoses and helps clinicians identify the precise nature of the language deficits. Furthermore, recognizing distinct disorder features allows for the development of targeted intervention strategies, enabling clinicians to design personalized and effective treatments that address the unique difficulties faced by each patient.

Utilizing ChatGPT to identify and classify language disorder features via a structured multi-label classification approach significantly refines the diagnostic capabilities in ASD assessments. This methodology not only enhances the accuracy of the diagnoses but also deepens the understanding of the patient's specific communicative deficits, facilitating the development of targeted therapeutic strategies.

\section{Conclusion}\label{sec13}

This research confirmed the substantial benefits of integrating LLMs such as ChatGPT with the ADOS-2 procedures for diagnosing ASD in adults. Utilizing ChatGPT, enhanced with Google's speaker diarization and transcription technologies, significantly improved the accuracy, PPV, sensitivity, and F1 score of language deficit diagnoses compared to traditional models. This integration streamlines the diagnostic process, improving efficiency and reducing subjectivity. It also enhances the scalability of interventions, enabling enhanced efficiency and precision assessments essential for effective treatment planning.

Looking forward, the study highlights the potential for these technologies to incorporate a wider variety of data and to develop adaptive learning models that continually improve in accuracy and effectiveness. This progression promises to revolutionize ASD diagnostics, paving the way for more personalized and accessible care for individuals with ASD. The integration of LLMs like ChatGPT in clinical settings is a forward step in making ASD diagnostics not only quicker and more accurate but also more comprehensive in understanding and addressing the diverse needs of the autism community.

\backmatter








\section*{Declarations}


\begin{itemize}
\item Funding

This material is based upon work supported by the National Science Foundation under Grant No. HCC-2401748 and National Institute of Health under Grant No. R01MH129426. The funders had no role in study design, data collection and analysis, decision to publish, or preparation of the manuscript.

\item Conflict of interest/Competing interests (check journal-specific guidelines for which heading to use)

The authors declare no conflict of interest. 
\item Author contribution

C.H., L.K.P., S.W., X.L. designed research. L.K.P. performed experiments. C.H., W.L., M.R., and X.Y. analyzed data. C.H., S.W., and X.L. wrote the paper. All authors discussed the results and contributed toward the manuscript.
\end{itemize}

\noindent

\bibliography{sn-bibliography}

\end{document}